\documentclass[letterpaper]{article} 
\usepackage{aaai25}  
\usepackage{times}  
\usepackage{helvet}  
\usepackage{courier}  
\usepackage[hyphens]{url}  
\usepackage{graphicx} 
\usepackage{natbib}  
\usepackage{caption} 
\usepackage{algorithm}
\usepackage{algorithmic}
\usepackage{amsmath}
\usepackage{adjustbox}
\usepackage{subfigure}
\usepackage{booktabs}
\usepackage{amssymb}
\usepackage{bm}
\usepackage{placeins}
\usepackage{xcolor}

\usepackage{newfloat}
\usepackage{listings}
\DeclareCaptionStyle{ruled}{labelfont=normalfont,labelsep=colon,strut=off} 
\floatstyle{ruled}
\newfloat{listing}{tb}{lst}{}
\floatname{listing}{Listing}
\newcounter{promptcount}

\newenvironment{prompt}{%
    \refstepcounter{promptcount}
    \par\medskip\noindent
    \minipage{\linewidth}
}{%
    \endminipage\par\medskip
}
\usepackage[multiple]{footmisc}
\title{Improving Complex Reasoning over Knowledge Graph\\ with Logic-Aware Curriculum Tuning}
\author{
    Tianle Xia\textsuperscript{\rm 1}, Liang Ding\textsuperscript{\rm 2}, Guojia Wan\textsuperscript{\rm 1}\thanks{Corresponding author.}, Yibing Zhan\textsuperscript{\rm 3}, Bo Du\textsuperscript{\rm 1}\footnotemark[1], Dacheng Tao\textsuperscript{\rm 4}
}
\affiliations{
    \textsuperscript{\rm 1}School of Computer Science, National Engineering Research
Center for Multimedia Software, \\Institute of Artificial Intelligence,
and Hubei Key Laboratory of Multimedia and Network Communication Engineering,\\ Wuhan University, China\\

    \textsuperscript{\rm 2} The University of Sydney\\
    \textsuperscript{\rm 3} JD Explore Academy\\
    \textsuperscript{\rm 4}College of Computing and Data Science at Nanyang Technological University, Singapore 639798\\
    \{2018302020065, guojiawan, dubo\}@whu.edu.cn,\{liangding.liam, dacheng.tao\}@gmail.com,zhanyibing@jd.com
}

\usepackage{bibentry}

\begin{document}

\maketitle

\begin{abstract}
Answering complex queries over incomplete knowledge graphs (KGs) is a challenging task. Most previous works have focused on learning entity/relation embeddings and simulating first-order logic operators with various neural networks. However, they are bottlenecked by the inability to share world knowledge to improve logical reasoning, thus resulting in suboptimal performance. In this paper, we propose a complex reasoning schema over KG upon large language models (LLMs), containing a curriculum-based logical-aware instruction tuning framework, named LACT. Specifically, we augment the arbitrary first-order logical queries via binary tree decomposition, to stimulate the reasoning capability of LLMs. To address the difficulty gap among different types of complex queries, we design a simple and flexible logic-aware curriculum learning framework. Experiments across widely used datasets demonstrate that LACT has substantial improvements~(brings an average +5.5\% MRR score) over advanced methods, achieving the new state-of-the-art. 
\end{abstract}
\begin{links}
    \link{Code}{https://github.com/TianleXia0914/LACT}
\end{links}

%

\section{Introduction}

Industrial-scale knowledge graphs~(KGs) like FreeBase store structural knowledge in a collection of fact triplets and are widely adopted by many domains. Unfortunately, KGs are often incomplete, leaving many missing triplets undiscovered. Thus, complex logical reasoning over such incomplete KGs~\cite{hamilton2018embedding,bai2023answering} is a challenging task and has attracted much attention in the recent years.
In the realm of complex logical queries, Existential First Order queries with a single free variable ($\bm{EFO_{1}}$) emerge as a powerful tool that includes logical operators such as the basic projection operation $(\exists)$ and the corresponding negation operation $(\neg)$, two relational operation intersection $(\wedge)$ and union $(\vee)$, etc.
A very straightforward approach is to represent the computation graph as a Directed Acyclic Graph (DAG), which can be addressed by a systematic traversal of a knowledge graph (KG). This process entails the allocation of suitable entities to intermediate variables based on their structural attributes~\cite{Dalvi_Suciu_2007}.
\begin{figure*}
    \centering
    \includegraphics[width=1\textwidth]{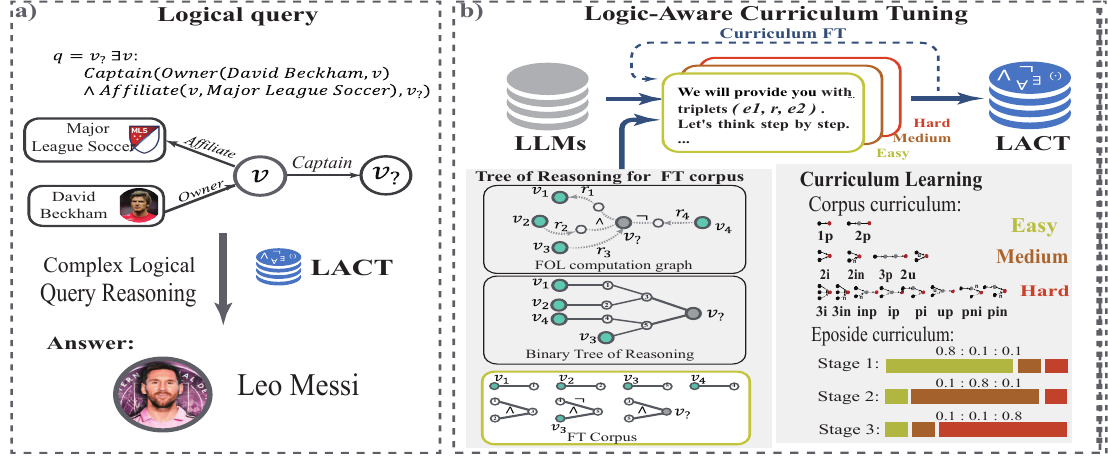}
    \caption{Schematic illustration: a) Answering logical query over KG with LACT. b) The framework of Logic-Aware Curriculum Tuning over LLama. We leverage the binary tree decomposition strategy (Seen in Section Methodology) to construct a logic-rich FT corpus and the Curriculum learning strategy (Seen in Section Methodology) to fine-tune a base LLM. c) Performing reasoning using well-designed prompts.}
    \label{fig:btm}
\end{figure*}

In the wake of knowledge graph embedding (KGE)'s resounding success,~\cite{Bordes_Usunier_Garcia-Duran_Weston_Yakhnenko_2013,Bai_Ying_Ren_Leskovec}, a series of work seeks to answer complex logical queries by learning query embedding and simulating logical operators with well-designed neural networks ~\cite{Chen_Hu_Sun_2021,Zhu_Galkin_Zhang_Tang_2022,Zhang_Wang_Chen_Ji_Wu_2021,Arakelyan_Daza_Minervini_Cochez_2020}.
In contemporary research, the exploration of KGE method predominantly centers around the intricate design of various embedding representations, including geometric embedding like Euclidean geometry modeling~\cite{hamilton2018embedding,Ren_Hu_Leskovec_2020} and Riemannian geometry modeling~\cite{Choudhary_Rao_Katariya_Subbian_Reddy_2021}, and embedding based on probability modeling~\cite{Ren_Hu_Leskovec_2020}, seeking to encapsulate the semantic representation and vector mapping of KGs' entities and relations.

However, the embedding-based approach described above also has some limitations. (1) \textbf{Limited information}: Knowledge graphs often contain incomplete and insufficient information. When only the information from the knowledge graph is available, it becomes challenging to perform complex reasoning that requires additional or missing information.
(2) \textbf{Insufficient utilization of semantic information}: Current knowledge graph embedding (KGE) methods primarily focus on the structural information of the knowledge graph, paying little attention to the actual semantics of entities and relationships.
(3) \textbf{High complexity of logical queries}: he intricate nature of real-world knowledge contributes to the complexity of reasoning in practical applications. As a result, it is challenging to accurately model the relationships within real-world knowledge using simple geometric representations. Such simplifications may overlook potentially complex relational information~\cite{Choudhary_Rao_Katariya_Subbian_Reddy_2021}, thereby limiting the effectiveness of advanced logical reasoning.
(4) \textbf{Generalizability}: A Knowledge Graph Embedding (KGE) model tailored to a specific knowledge graph cannot generalize across different knowledge graphs. This limitation hinders the practical application of such approaches in real-world scenarios, where knowledge graphs often vary significantly in both structure and content.

Recently, large language models (LLMs)
~\cite{OpenAI_2023,Touvron_Lavril_Izacard_Martinet_Lachaux_Lacroix_Rozi`ere_Goyal_Hambro_Azhar_et_al.} have showed outperforming capabilities for a wide range of tasks~\cite{Ouyang_Wu_Jiang_Almeida_Wainwright_Mishkin_Zhang_Agarwal_Slama_Ray_etal.,zhong2023can,peng2023towards,lu2023error,guo2025multi}. With the rise of this trend,~\cite{choudhary2023complex,liu2024logic} construct prompt templates and apply LLMs as text-generators to answer complex queries. However, LLM without fine-tuning suffers from hallucination problem~\cite{Zhang_Li_Cui_Cai_Liu_Fu_Huang_Zhao_Zhang_Chen_et_al._2023}, especially when faced with such a knowledge-intensive task that generates answers on an incomplete KG rather than simple retrieval. Besides, previous tasks relied on manual classification of queries to improve performance, which is unrealistic in large-scale practical applications and also limits the types of queries. Finally, Previous methods generally decompose the problem into sub-problems, which greatly increases the \textbf{cost} of reasoning, especially considering that previous methods are generally based on closed-source models such as GPT. Therefore, compared to pure prompt engineering, we prefer to fine-tune our model to solve the above problems.

In this paper, we propose \textbf{L}ogic-\textbf{A}ware \textbf{C}urriculum \textbf{T}uning (\textbf{LACT}\includegraphics[width=3.5 mm]{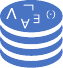}), a novel fine-tune framework for answering complex logical query, which stimulates the ability of LLMs to perform complex reasoning on knowledge graphs. We propose a strategy to incorporate the knowledge contained in the KGs into our training corpus to activate the corresponding knowledge of the LLMs and supplement the missing relevant knowledge of the LLMs during the fine-tuning process. At the same time, we have proven that data argument by binary tree decomposition can arouse the corresponding capabilities of LLMs and effectively improve their reasoning performance. At last, we show that curriculum learning~\cite{bengio2009curriculum} can effectively smooth the difficulty differences between different types of queries and greatly improve the results of difficult queries. In summary, our contribution manifests in three distinct facets:

\begin{itemize}
    \item We propose a \textbf{L}ogic-\textbf{A}ware \textbf{C}urriculum \textbf{t}uning (\textbf{LACT}) paradigm for complex logical reasoning over KGs.
    \item \textbf{LACT} achieves state-of-the-art performance beyond embedding-based and PLM-based methods, using only 7B models.
    \item Through extensive experimentation, we found that SFT Corpus constructed by queries on incomplete KGs via \textbf{Binary Tree Decomposition} and \textbf{Curriculum Learning} can significantly enhance LLM logical reasoning ability.
\end{itemize}

\section{Related Works}

\subsection{Logical Reasoning over Knowledge Graph}

Knowledge graphs can be widely used to enhance natural language reasoning~\cite{zhang2022knowledge,zhao2023ke}, understanding~\cite{zhong2023knowledge,liu2023unified}, and generation~\cite{ding2023recurrent,pan2024pomp}.
Given a $EFO_{1}$ query over a KG, complex logical reasoning aims to answer the correct entities, which contains both multi-hop queries and logical operators\cite{Guu_Miller_Liang_2015,hamilton2018embedding} over a incomplete KG. Most of the current approaches have focused on learning meaningful KG embeddings including \cite{Arakelyan_Daza_Minervini_Cochez_2020,Chen_Hu_Sun_2021,Zhang_Wang_Chen_Ji_Wu_2021,Zhu_Galkin_Zhang_Tang_2022,wang2023query}. Neuralizing logical operators through a specific embedding space, thereby embedding $EFO_{1}$ queries into a multi-dimensional vector space\cite{hamilton2018embedding,Ren_Hu_Leskovec_2020}, or into a specific probability distribution\cite{Ren_Hu_Leskovec_2020,Choudhary_Rao_Katariya_Subbian_Reddy_2021}, then obtain the final answer set by fitting the nearest neighbour representation based on relational operations. 
Additionally, recent work like CQD~\cite{Arakelyan_Daza_Minervini_Cochez_2020} its optimization work~\cite{arakelyan2024adapting} improves performance by working to reduce query difficulty by decomposing complex queries into simple one-hop queries, QTO~\cite{bai2023answering} introduces query computation tree optimization and LMPNN~\cite{wang2023logical} aggregates logical messages passed on the query graph by training an MLP network. Some work~\cite{yin2023rethinking} has begun to expand into the area of cyclic query. Despite their efficacy, embedding-based methods often suffer from a lack of generalization due to their specificity in embedding knowledge graphs. Meanwhile, this limitation on the embedding range hinders their ability to effectively generalize to more complex query structures.

Moreover, PLM-based methods view complex logical reasoning as text-generation tasks by modeling pre-trained language models. $EFO_{1}$ queries are organized into input-output sequence pairs after textualization and encoded by PLM~\cite{wang2023unifying,xu2023prediction,wang2023unifying}. However, limited by the performance limitations of the base model, the PLM method has always been deficient in understanding world knowledge and reasoning capabilities, limiting its performance in complex reasoning.
\subsection{LLMs for KG Reasoning}
In recent years, substantial advancements have been witnessed in the domain of LLMs \cite{OpenAI_2023,touvron2023llama,peng2023towards,zhong2023can,Zeng_Liu_Du_Wang_Lai_Ding_Yang_Xu_Zheng_Xia_et_al._2022}. Among these, instruction tuning (IT)~\cite{Ouyang_Wu_Jiang_Almeida_Wainwright_Mishkin_Zhang_Agarwal_Slama_Ray_etal.} and the alignment of the model~\cite{Wang_Zhong_Li_Mi_Zeng_Huang_Shang_Jiang_Liu} with human preferences stand out. 

Within the realm of LLM, the integration of LLMs with Knowledge Graphs (KG)~\cite{Pan_Luo_Wang_Chen_Wang_Wu_2023,Wang_Zhao_Qiang_Li_Xi_Du_Cai_Guo_Chen_Xu_et_al._2023,luo2023reasoning} constitutes a prominent and consequential research avenue. 

Leveraging its potent generative capabilities, LLMs prove invaluable in addressing Knowledge Graph-related tasks, including but not limited to Knowledge Graph Completion (KGC)~\cite{Zhu_Wang_Chen_Qiao_Ou_Yao_Deng_Chen_Zhang,zhang2023making}, entity alignment~\cite{Zhang_Su_Trisedya_Zhao_Yang_Cheng_Qi_2023}, Knowledge Graph Question Answering (KGQA)~\cite{luo2023reasoning}, and others~\cite{luo2023chatrule}. Consequently, the synergy between Knowledge Graphs for LLMs (KG4LLM) and LLMs for Knowledge Graphs (LLM4KG) emerges as an essential focal point, bearing significance in advancing the collective capabilities of both entities.

Our research endeavours center around the utilization of Language Models (LLMs) for the Complex Logical Reasoning task, an area that remains relatively unexplored. \cite{choudhary2023complex} made the initial attempt by prompt engineer but it lacks in-depth research and simply uses LLMs as text generators.
\subsection{Curriculum Learning}
The concept of progressively training neural networks from simple to complex configurations has its origins dating back to~\cite{Elman_1993,Krueger_Dayan_2009}. Based on these works, Curriculum Learning is first proposed in~\cite{bengio2009curriculum}. A series of illustrative experiments were meticulously crafted to showcase the advantages of employing curriculum strategies in both image classification and language modeling. When focusing on the field of NLP, by experimenting with several heuristics, ~\cite{Sachan_Xing_2016,Xu_Zhang_Mao_Wang_Xie_Zhang_2020} migrated the success of CL to NLU tasks. \cite{ding2021progressive,zhou2021self}
 improved the machine translation modeling by carefully designing different curricula.
 Recently, with the rise of LLM, \cite{liu2024let} discovered the huge potential of CL in in-context learning, while \cite{wang2023gradual} focus on the improvement of CL for LLM in pre-train. However, we are committed to exploring the huge potential of CL in fine-tuning LLM.
\section{Preliminary}
\subsection{Knowledge Graph}
In our work, a knowledge graph is $\mathcal{G} = (\mathcal{E,R,T})$ where $\mathcal{E,R}$ are the set of entity and relation respectively. With regard to generality, KG can be considered as a collection of triplets $\{\mathcal{T} = (v_s,r,v_t) | v_s,v_t \in \mathcal{E}, r \in \mathcal{R}\}$, where $v_s/v_t$ denotes the head/tail entity.
\subsection{Complex logical query}
Complex logical query is used for retrieving and manipulating data stored in knowledge graphs, which is grounded in a subset of $EFO_{1}$ query. The process of answering a complex logical query involves trying to match suitable results using the composition of queries:
\begin{equation}
	q[v_?]=\exists v:q_1\wedge q_2\wedge\cdots\wedge q_n,
	\label{NF}
\end{equation}
or,
\begin{equation}
	q[v_?]=\exists v:q'_1\vee q'_2\vee\cdots\vee q'_n,
	\label{DNF}
\end{equation}
where $q$ denotes a $EFO_{1}$ query and $q_i$ denotes a sub-query. Note that Eq. (\ref{NF}) is a conjunctive normal form (CNF) and Eq. (\ref{DNF}) is a disjunctive normal form (DNF) \cite{Ren_Hu_Leskovec_2020}. The two can be equivalently converted to each other via De Morgan's law. 


\section{Methodology}
\subsection{Instruction Tuning on LLMs}

Within this section, we elucidate the methodology for seamlessly integrating Knowledge Graph (KG) information into the textual prompt.

In the context of complex logical reasoning, we designate a Language Model (LLM) as $\mathcal{M}$ serving as a text decoder to produce the corresponding output. By commencing with the aforementioned definition, we can frame this task as a text generation problem.
n contrast to vanilla text generation, triplet generation involves more complexity due to the intricate semantic information associated with the entities and relations in the triplet prompt, as defined by the given knowledge graph (KG). 
In fact, we want the generated answers to be entities that exist in KG itself.
Without this knowledge, the predicted answers are unreliable and unstable. Thus, to engage LLM in complex logical reasoning, a crucial step involves incorporating knowledge graph (KG) information into the prompt. This integration provides additional auxiliary context, effectively enhancing the performance of LLM.
                                   
In particular, when we fine-tune $\mathcal{M}$, we can treat the training corpus as a set of question-answer pairs $(\mathcal{S}, \mathcal{A})$. When considering complex reasoning over KGs, the input textual sequence $\mathcal{S}$ consists of the description of question $\mathcal D$, knowledge graph neighbourhood information(i.e. related triplets) $\mathcal{X}$ and logical query. In our work, we used a simple but effective method called \textit{greedy depth traversal algorithm} to search for neighbourhood information and we simply discarded all samples that exceeded the token limit. (Detailed algorithm and token distribution can be found in Appendix). The logical query contains the textual information about the query $q_{\tau}$ with the query structure $\tau$ and specific query content $Q_{\tau}$ that needs to be processed, which can be denoted as $f_{1}(q_{\tau})$. Likewise,we can denote the output answer A as $f_{2}(V_{\tau})$, where $f_{2}$ indicates textualization of $V_{\tau}$ here. In summary, the fine-tune training corpus $\mathcal C$ can be expressed in the following form:
\begin{equation}
    \mathcal C=(\mathcal{S},\mathcal{A})=(\mathcal D \oplus \mathcal X \oplus f_{1}(q_{\tau}),f_{2}(V_{\tau})).
\end{equation}

The model $\mathcal{M}$(parameterized by $\theta$) is fine-tuned by the next token prediction task. We fine-tune $\mathcal{M}$ to obtain the final model by maximizing the log-likelihood of the next token. The training loss can be formulated as
\begin{equation}
\mathcal{L}=-\frac{1}{\left|\mathcal{C}\right|} \sum_{i=1}^{\left|\mathcal{C}\right|} \log P_{\mathcal{M}}\left(c_{i} \mid c_{<i}\right),
\label{eq4}
\end{equation}
where $c_i(i=1,2,...,\left|\mathcal{C}\right|)$ represents the textual tokens of the training corpus $\mathcal{C}$. For our task, the training objective can be transferred as
\begin{equation}
    \mathcal{L}=-\frac{1}{\left|\mathcal{C}\right|} \sum_{i=1}^{\left|\mathcal{C}\right|} \log P_{\mathcal{M}}\left(\mathcal{A} \mid \mathcal{S}\right).
\end{equation}
\subsection{Data Augmentation by Binary Tree Decomposition}
\label{binarytree}
This section introduces how to build fine-tuning corpora that make LLMs (Large Language Models) logic-aware based on instruction tuning.

Chain of thought (COT) enables models to decompose multi-step problems into intermediate steps, subsequently improving the reasoning abilities of LLMs~\cite{wei2022chain}. However, pure-prompt-based reasoning needs more in-context memory to perform complex logical reasoning. Considering complex logical queries, which query structure can be transferred into the form of a DAG and its hierarchical structure becomes a natural fit for decomposition into a series of sub-problems. So we propose a method for data augment based on \textit{Binary Tree Decomposition Mechanism} to stimulate LLMs with the potential to decompose a complex query into a chain of simple queries.
\noindent\textbf{Binary Tree Decomposition Mechanism.} The Binary Tree Decomposition Mechanism is divided into the following three steps:

\textbf{Query Computation Tree.} Existing embedding-based methods rely on DNF to avoid embedding divergence \cite{Ren_Leskovec_2020}. Our method bypasses logical operator learning. Consequently, this allows us to convert any query into a logic-equivalent tree form through iterative sub-query recursion. For a complex $EFO_{1}$ query, like the example shown in Figure \ref{fig:btm}, its computation graph that is a directed acyclic graph can be converted into a tree where the root node is $v_?$. The tail entities and intermediate entities in complex queries are in one-to-one correspondence with the root nodes and leaf nodes in the generated calculation tree respectively. In view of the reverse order that the tail entity brings to the root node, the edges in the tree corresponding to the relationship in each query are child-to-parent directions. In the Appendix, We demonstrate a simple but systematic method for converting complex queries into computational trees.

\textbf{Binary Tree Decomposition.}
We split each 1-to-n intersection/union node into n corresponding child nodes.
Consider that merging union branches may result in a 1-to-n structure consisting of intersection and union edges, as shown in Figure~\ref{fig:btm}.
This can be properly addressed by separating $v_?$ into an intersection node structure ($v_3'$ and $v_5'$ in Figure~\ref{fig:btm}) firstly, then decomposing the node that was decomposed in the previous layer into a deeper intersection node structure ($v_1'$ and $v_2'$ for $v_3'$, $v_4', v_3$ for $v_5'$), where $v'$ denotes an intermediate entity retrieved by Neighborhood Retrieval Algorithm (Seen in Appendix).

\textbf{Reverse Level Traversal.} Finally, we decompose the binary computation tree into independent branches. Since the root node of the calculation tree is the answer entity, we perform a hierarchical traversal of all non-leaf nodes of the binary tree in reverse. As shown in Figure~\ref{fig:btm}, the complex $EFO_{1}$ query is decomposed into a sequence:
\begin{align*}
    [(v_1,r,v_1')&,(v_2,r,v_2'),(v_3,r,v_5'),(v_4,r,v_4'),\\
    (v_1',r,v_3')&,\wedge,(v_2',r,v_3'),\neg,(v_4',r,v_5'),\wedge,(v_3,r,v_5'),\\
    (v_3',r,v_?)&,\wedge,(v_5',r,v_?)].
\end{align*}

\noindent\textbf{Data Augmentation.} Now we can turn any loopless $EFO_{1}$ query into a series of separate subqueries. We use a defined template to integrate the decomposition process into the answers to the training corpus.
So, the training corpus $\mathcal C$ can be transferred into the following form:
\begin{equation}
    \mathcal C=(\mathcal{S},\mathcal{A})=(\mathcal D \oplus \mathcal X \oplus f_{1}(q_{\tau}),f_{2}( V_{\tau,\mathcal{D}ecomposed})),
\end{equation}
where $V_{\tau,\mathcal{D}ecomposed}$ indicates the answer corresponding to the logical query with the decomposition reasoning path.
\subsection{Fine-tuning Enhanced by Curriculum Learning}
\label{curriculum}
As mentioned in previous sections, though decomposing into chain responses, complex queries still vary greatly in difficulty and complexity due to differences between query structures.

Naturally, we believe that these different types of samples should not be simply lumped together. Intuitively, we incorporate curriculum learning into our training. To be specific, given the particularity of complex reasoning data, when we decompose it into logical chains, naturally, we can use the number of decomposed sub-logical queries as a natural difficulty discriminator to select different types of queries, e.g., a 1p query would be defined as difficulty-1, while a 2p query, which can be decomposed into two projection queries and an intersection query, would be defined as difficulty-3. The detailed difficulty discriminating process will be shown in the Appendix.

Finally, we divided samples into three parts: easy samples, medium samples and difficult samples according to the difficulty level. Correspondingly, our training process is also divided into three stages. After we did some exploratory experiments, we did not simply train three data sets in the order of easy-medium-difficult. On the contrary, we decided to first use 80\% easy samples, 10\% medium samples, and 10\% difficult samples for the first stage of training and the subsequent two-stage training process is a Leto, and experimental results in the next few sections also proved that this is effective.
\begin{table*}[t]
    \centering
    \begin{adjustbox}{width=\textwidth}
    \begin{tabular}{lccccccccccccccccc}
        \toprule
         Method & $avg_{p}$ & $avg_{ood}$ & $avg_{n}$ & 1p & 2p & 3p & 2i & 3i & pi & ip & 2u & up & 2in & 3in & inp & pin & pni \\
         \midrule
         \textbf{Dataset}& \multicolumn{17}{c}{FB15K}\\
         \midrule
         \textbf{GQE}&28.0 & 20.1 & - & 54.6 & 15.3 & 10.8 & 39.7 & 51.4 & 27.6 & 19.1 & 22.1 & 11.6 & - & - & - & - & -\\
         \textbf{Query2Box} & 38.0 & 29.3 & - & 68.0 & 21.0 & 14.2 & 55.1 & 66.5 & 39.4 & 26.1 & 35.1 & 16.7 & - & - & - & - & -\\
         \textbf{BetaE }& 41.6 & 34.3 & 11.8 & 65.1 & 25.7 & 24.7 & 55.8 & 66.5 & 43.9 & 28.1 & 40.1 & 25.2 & 14.3 & 14.7 & 11.5 & 6.5 & 12.4 \\
        \textbf{CQD-CO} & 46.9 & 35.3 & - & 89.2 & 25.3 & 13.4 & 74.4 & 78.3 & 44.1 & 33.2 & 41.8 & 21.9 & - & - & - & - & - \\
        \textbf{CQD-Beam} & 58.2 & 49.8 & - & 89.2 & 54.3 & 28.6 & 74.4 & 78.3 & 58.2 & 67.7 & 42.4 & 30.9 & - & - & - & - & - \\
        \textbf{ConE} & 49.8 & 43.4 & 14.8 & 73.3 & 33.8 & 29.2 & 64.4 & 73.7 & 50.9 & 35.7 & 55.7 & 31.4 & 17.9 & 18.7 & 12.5 & 9.8 & 15.1 \\
        \textbf{GNN-QE} & 72.8 & 68.9 & 38.6 & 88.5 & 69.3 & 58.7 & 79.7 & 83.5 & 69.9 & 70.4 & 74.1 & 61.0 & 44.7 & 41.7 & 42.0 & 30.1 & 34.3 \\
        \textbf{QTO}& 74.0 & 71.8 & 49.2 & 89.5 & 67.4 & 58.8 & 80.3 & \textbf{83.6} & 75.2 & 74.0 & \textbf{76.7} & \textbf{61.3} & 61.1& 61.2 & 47.6 &\textbf{48.9} & 27.5 \\
        \textbf{LARK} &56.1 & 43.1 & 18.4  & 72.8 & 50.7 & 36.2 & 66.9 & 60.4 & 56.1 & 23.5 & 52.4 & 40.6& 16.2 & 5.7 &33.7 & 26.1& 10.0\\
        \midrule
        \textbf{LACT} &\textbf{82.6} & \textbf{71.9}  & \textbf{56.9}&\textbf{93.5} & \textbf{73.5} & \textbf{59.6} & \textbf{92.3} & 82.3 &\textbf{76.8} &\textbf{75.9} & 74.6& 60.4 & \textbf{81.2} & \textbf{61.6} & \textbf{52.0}&43.5&\textbf{41.7}\\
        \midrule
        \textbf{Dataset} &\multicolumn{17}{c}{FB15K-237}\\
        \midrule
        \textbf{GQE} & 16.3 & 10.3 & - & 35.0 & 7.2 & 5.3 & 23.3 & 34.6 & 16.5 & 10.7 & 8.2 & 5.7 & - & - & - & - & - \\
        \textbf{Query2Box} & 20.1 & 15.7 & - & 40.6 & 9.4 & 6.8 & 29.5 & 42.3 & 21.2 & 12.6 & 11.3 & 7.6 & - & - & - & - & - \\
        \textbf{BetaE} & 20.9 & 14.3 & 5.5 & 39.0 & 10.9 & 10.0 & 28.8 & 42.5 & 22.4 & 12.6 & 12.4 & 9.7 & 5.1 & 7.9 & 7.4 & 3.5 & 3.4 \\
        \textbf{CQD-CO} & 21.8 & 15.6 & - & 46.7 & 9.5 & 6.3 & 31.2 & 40.6 & 23.6 & 16.0 & 14.5 & 8.2 & - & - & - & - & - \\
        \textbf{CQD-Beam} & 22.3 & 15.7 & - & 46.7 & 11.6 & 8.0 & 31.2 & 40.6 & 21.2 & 18.7 & 14.6 & 8.4 & - & - & - & - & - \\
        \textbf{FuzzQE} & 24.0 & 17.4 & 7.8 & 42.8 & 12.9 & 10.3 & 33.3 & 46.9 & 26.9 & 17.8 & 14.6 & 10.3 & 8.5 & 11.6 & 7.8 & 5.2 & 5.8 \\
        \textbf{ConE} & 23.4 & 16.2 & 5.9 & 41.8 & 12.8 & 11.0 & 32.6 & 47.3 & 25.5 & 14.0 & 14.5 & 10.8 & 5.4 & 8.6 & 7.8 & 4.0 & 3.6 \\
        \textbf{GNN-QE} & 26.8 & 19.9 & 10.2 & 42.8 & 14.7 & 11.8 & 38.3 & 54.1 & 31.1 & 18.9 & 16.2 & 13.4 & 10.0 & 16.8 & 9.3 & 7.2 & {7.8} \\
        \textbf{QTO} & 33.5 & 27.6 & 15.5 & 49.0 & 21.4 & 21.2 & 43.1 & 56.8 & 38.1 & 28.0 & 22.7 & 21.4 & 16.8 & 26.7 & 15.1 & 13.6 & 5.4 \\
        \textbf{LARK} & 50.7 & 41.0 & 10.6 & 73.6 & 40.5 & 26.8 & 46.1 & 43.1 & 49.9 & 22.9 & \textbf{62.8} & 28.3& 6.5 & 3.4 &23.2 & 16.5& 3.2\\
        \midrule
        \textbf{LACT} &\textbf{57.0} & \textbf{44.4} & \textbf{21.9}& \textbf{76.5} & \textbf{54.3} & \textbf{30.3} & \textbf{56.0} & \textbf{54.5} & \textbf{54.6} &\textbf{36.9} &56.5 &\textbf{29.7}&\textbf{17.6}&\textbf{33.1}&\textbf{27.1}&\textbf{19.8}&\textbf{11.2}\\
        \midrule
        \textbf{Dataset} &\multicolumn{17}{c}{NELL995}\\
        \midrule
        \textbf{GQE} & 18.6 & 12.5 & - & 32.8 & 11.9 & 9.6 & 27.5 & 35.2 & 18.4 & 14.4 & 8.5 & 8.8 & - & - & - & - & - \\
        \textbf{Query2Box} & 22.9 & 15.2 & - & 42.2 & 14.0 & 11.2 & 33.3 & 44.5 & 22.4 & 16.8 & 11.3 & 10.3 & - & - & - & - & - \\
        \textbf{BetaE} & 24.6 & 14.8 & 5.9 & 53.0 & 13.0 & 11.4 & 37.6 & 47.5 & 24.1 & 14.3 & 12.2 & 8.5 & 5.1 & 7.8 & 10.0 & 3.1 & 3.5 \\
        \textbf{CQD-CO} & 28.8 & 20.7 & - & 60.4 & 17.8 & 12.7 & 39.3 & 46.6 & 30.1 & 22.0 & 17.3 & 13.2 & - & - & - & - & - \\
        \textbf{CQD-Beam} & 28.6 & 19.8 & - & 60.4 & 20.6 & 11.6 & 39.3 & 46.6 & 25.4 & 23.9 & 17.5 & 12.2 & - & - & - & - & - \\
        \textbf{FuzzQE} & 27.0 & 18.4 & 7.8 & 47.4 & 17.2 & 14.6 & 39.5 & 49.2 & 26.2 & 20.6 & 15.3 & 12.6 & 7.8 & 9.8 & 11.1 & 4.9 & 5.5 \\
        \textbf{ConE} & 27.2 & 17.6 & 6.4 & 53.1 & 16.1 & 13.9 & 40.0 & 50.8 & 26.3 & 17.5 & 15.3 & 11.3 & 5.7 & 8.1 & 10.8 & 3.5 & 3.9 \\
        \textbf{GNN-QE} & 28.9 & 19.6 & 9.7 & 53.3 & 18.9 & 14.9 & 42.4 & 52.5 & 30.8 & 18.9 & 15.9 & 12.6 & 9.9 & 14.6 & 11.4 & 6.3 & 6.3 \\
        \textbf{QTO} & 32.9 & 24.0 & 12.9 & 60.7 & 24.1 & 21.6 & 42.5 & 50.6 & 31.3 & 26.5 & 20.4 & 17.9 & 13.8& 17.9& 16.9& 9.9& 5.9 \\
        \textbf{LARK} & 52.9 & 26.9 & 12.4 & 87.8 & 45.7 & 33.5 & 51.3 & 48.7 & 23.1 & 22.2 & 20.6 & 41.1& 9.9 & 5.9 &24.5 & 13.3& 7.3\\
        \midrule
        \textbf{LACT}& \textbf{60.1} & \textbf{32.0} & \textbf{17.2} & \textbf{91.4}&\textbf{53.6}&\textbf{40.6}&\textbf{62.2}&\textbf{54.9}&\textbf{31.4}&\textbf{34.8}&\textbf{27.0}&\textbf{34.0}&\textbf{16.0}&\textbf{21.2}&\textbf{21.0}&\textbf{16.3}&\textbf{11.6}\\ 
        \bottomrule
    \end{tabular}
    \end{adjustbox}
    \caption{\textbf{MRR} scores (\%) on complex reasoning for\textbf{ 14 types of queries}. $avg_{p}$ represents the mean score of nine ordinary queries including  1p/2p/3p/2i/3i and pi/ip/2u/ip; $avg_{ood}$ is the mean score of out of distribution (OOD) queries, which consist pi/ip/2u/ip queries; $avg_{n}$ is the mean score of queries including negation operation.}
    \label{tb:main}
\end{table*}
\subsection{Reasoning Module}
We retrieve relevant information and textualize the $EFO_{1}$ query, and finally, we populate it into the template in the Appendix to generate queries.

As shown in Appendix, we use the LLM to do a simple text generation task to get the answer. After fine-tuning, LACT can follow the output mode in the training stage in Figure~\ref{fig:btm}, so we can extract final answers through simple regular expressions with the template in the Appendix. The specific prompt can be found in the Appendix.

\section{Experiments}
\subsection{Training Settings}

We opt for the most popular datasets: \textbf{FB15K}, \textbf{FB15K-237}, \textbf{NELL995}. We used the training set of the above dataset as the original training data. Detailed information about the dataset and training set is listed in the Appendix.

\subsection{Experimental Settings}
\noindent \textbf{Baseline Methods} For comparing with KGE, we chose the following representative methods as baselines:
\textbf{GQE}~\cite{hamilton2018embedding},
\textbf{Query2Box(Q2B)}~\cite{choudhary2023complex},
\textbf{BetaE}~\cite{Ren_Leskovec_2020},
\textbf{CQD}~\cite{Arakelyan_Daza_Minervini_Cochez_2020},
\textbf{ConE}~\cite{Zhang_Wang_Chen_Ji_Wu_2021},
\textbf{GNN-QE}~\cite{Zhu_Galkin_Zhang_Tang_2022},
\textbf{QTO}~\cite{bai2023answering}.
Moreover, we also compared our LACT with the LLM-based method \textbf{LARK} ~\cite{choudhary2023complex}.
\begin{figure}[htbp]
    \centering
    \includegraphics[width=1\linewidth]{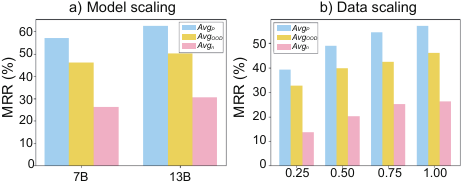}
    \caption{Performance of \textbf{scaling} LACT on FB15K-237 with different a) \textbf{model} and b)\textbf{ data scales}.}
    \label{fig:sl}
\end{figure}
\begin{figure}[htbp]
	\centering
        \includegraphics[width=1\columnwidth]{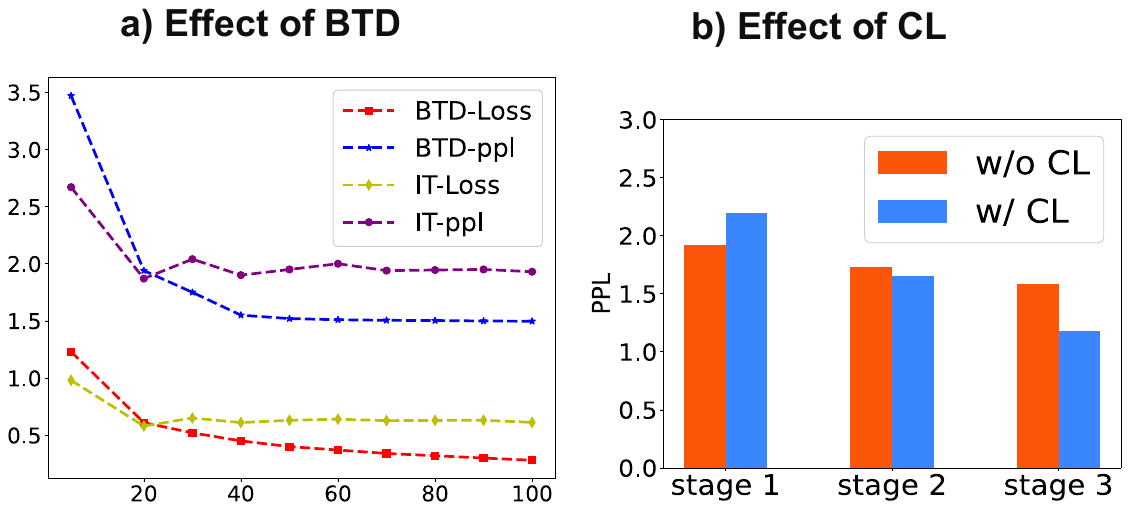}
        \caption{Results of \textbf{Ablation Studies.} (a) Comparison \textbf{PPL and train loss} results of whether to \textbf{use BTD} based on FB15k. (b) Comparison \textbf{PPL results} of whether use \textbf{CL} based on FB15K high-difficulty queries.}
        \label{fig:ab}
\end{figure}
\begin{table*}[htbp]
    \centering
    \begin{adjustbox}{width=1\textwidth}
    \begin{tabular}{lcccccccccccccc}
    \toprule
    \bf{Method}  & \bf{1p} & \bf{2p} & \bf{3p} & \bf{2i} & \bf{3i} & \bf{pi} & \bf{ip} & \bf{2u} & \bf{up} & \bf{2in} & \bf{3in} & \bf{inp} & \bf{pin} & \bf{pni} \\
    \midrule
    Llama2 &  67.2 & 42.3 & 38.3 & 61.6 & 44.8 & 34.1 & 36.9 & 44.2 & 28.4 & 44.7 & 38.5 & 36.9 & 32.1 & 30.0 \\
    \quad  + IT&  94.6 & 68.8 &60.2&84.5&66.7 & 56.0 & 60.2 & 69.5 & 42.3 & 66.5 & 54.4 & 41.0 & 38.8 & 36.9 \\
    \quad \quad  +BTD &  91.5 & 72.3 & 65.6 & 89.7 & 75.2 & 60.1 & 65.4 & 72.5 & 49.9 & 74.3 & 66.4 & 48.9 & 46.5 & 42.5  \\
    \toprule
    \end{tabular}
    \end{adjustbox}
    \caption{\textbf{Accuracy} results (\%) evaluated on \textbf{FB15k} of whether to \textbf{use BTD} for \emph{hard} complex query on all query types.}
    \label{tb:ab1}
\end{table*}

\noindent\textbf{Evaluation Protocol}
Following previous works \cite{Ren_Hu_Leskovec_2020}, we choose the Mean Reciprocal Rank (MRR) with standard evaluation protocol for evaluating complex reasoning on KG. The detailed setting can be found in the Appendix. 
\subsection{Main Results}
The main experiment results of three datasets are shown in Table~\ref{tb:main}. The baseline training datasets contain five different queries including 1p/2p/3p/2i/3i, so another four queries are considered as OOD types, so the mean value of OOD types of queries is recorded as $avg_{ood}$. In our experiments, LACT consistently outperforms baseline methods across all datasets. Notably, LACT yields an average gain of 7.3\%, 2.9\%, and 6.3\% on $avg_{p}$, $avg_{odd}$, and $avg_{n}$, compared to the previous SOTA method, especially more challenging datasets like FB15K-237 and NELL995. Our method exhibits superior reasoning capability and effectively captures a wide range of internal relations, leading to enhanced performance on complex queries.
\begin{table*}[ht]
    \centering
    \begin{adjustbox}{width=1\textwidth}
    \begin{tabular}{lcccccccccccccccc}
    \toprule
    &\bf{BTD} &\bf{CL}  & \bf{1p} & \bf{2p} & \bf{3p} & \bf{2i} & \bf{3i} & \bf{pi} & \bf{ip} & \bf{2u} & \bf{up} & \bf{2in} & \bf{3in} & \bf{inp} & \bf{pin} & \bf{pni} \\
    \midrule
    Llama2 w/IT & &   & 94.6 & 68.8 &60.2&84.5&66.7 & 56.0 & 60.2 & 69.5 & 42.3 & 66.5 & 54.4 & 41.0 & 38.8 & 36.9 \\
    \midrule
    &\checkmark&  & 94.2 & 72.3 & 65.6 & 89.7 & 75.2 & 60.1 & 65.4 & 72.5 & 49.9 & 74.3 & 66.4 & 48.9 & 46.5 & 42.5    \\
   & & \checkmark&   94.6 & 70.9 &61.3&84.4&72.7 & 58.2 & 63.2 & 69.5 & 47.3 & 68.5 & 64.0 & 42.3 & 40.9 & 37.1 \\
    &\checkmark&\checkmark   & 94.8 & 78.7 & 69.2 & 94.8 & 88.1 & 79.3 & 80.5 & 80.7 & 67.1 & 90.6 & 70.4 & 59.3 & 53.6 & 46.7  \\
    \toprule
    \end{tabular}
    \end{adjustbox}
    \caption{\textbf{Accuracy }results (\%) evaluated on \textbf{FB15k} of whether to \textbf{use CL} for \emph{hard} complex query on all query types.}
    \label{tb:ab2}
\end{table*}

\subsection{Ablation Studies}

To validate the efficacy of the LACT module, we conduct a two-part ablation study. Firstly, we will investigate the impact of logical chain decomposition, while the second part assesses the effectiveness of curriculum learning.”

\noindent\textbf{Effect of Binary Tree Decomposition (BTD).} As shown in Figure~\ref{fig:ab}, logical chain decomposition can stimulate LLM's ability of logical decomposition, thereby greatly improving the performance of difficult queries.  

From a training perspective, as shown in Figure~\ref{fig:ab}, although perplexity (PPL) and training loss of decomposed queries before training were slightly higher than that of ordinary queries, we found that as training progresses, the loss and PPL of decomposed queries will quickly decrease to levels much lower than ordinary queries, proving that chain decomposition is effective to reduce the difficulty of learning complex queries.

\noindent\textbf{Effect of Curriculum Learning.} Curriculum learning, as illustrated in Table~\ref{tb:ab2}, greatly alleviates the gap between difficult training samples and the understanding ability of LLMs.

We can observe from Figure~\ref{fig:ab} that compared with random shuffle sequence training, difficult training samples under curriculum learning gradually become easier to understand. It is worth mentioning that we found that the gain of curriculum learning on training corpus that has not been decomposed by logical chains is very small, which supports our theory from the side. It is difficult for LLMs to understand the difficulty difference between undecomposed samples, so curriculum learning is also difficult to take effect.

\subsection{Transferability Study}
Considering the diversity of complex reasoning tasks, we can divide transferability into two levels, task-level and dataset-level transferability.

\noindent\textbf{Task-level transferability.}
The results in Table~\ref{tb:main} show that our LACT achieves a relative gain of 9.9\% on the OOD task, which demonstrates the strong generalization of our fine-tuning framework. Even in the OOD queries, as shown in the Appendix, more than 95\% of test samples can still follow logical chain reasoning. These phenomena indicate the strong generalization ability of LACT.

\noindent\textbf{Dataset-level transferability.} In fact, almost all KGE methods, even if some of the optimization methods claim not to require training, require a KGE base model adapted to a specific dataset, which leads to the inherent defect of extremely poor Transferability of the KGE method. However, as previous research has shown, fine-tuning LLMs is mainly to stimulate the knowledge and capabilities of potentially relevant downstream tasks contained in LLM pre-training. This has also become the theoretical basis for the transferability of fine-tuning methods for LLMs. The results in the Appendix show that the reasoning ability stimulated by one dataset can still be demonstrated in another dataset, which reflects well in the query performance which only dropped less than 5\%.

\noindent\textbf{Model-level transferability.} We tried analytical experiments with different base models to determine whether our LACT was universal. Obviously, The results in the Appendix show that all types of queries have been improved to a certain extent due to the progress of the base model. Experimental results show that our LACT is suitable for different pedestal models and has strong generalization.
\subsection{Scalability Study}

To verify the scalability of LACT, we scale LACT to different model sizes and data volumes.

\noindent\textbf{Performance on different model size.} We tried scaling model size to see if LACT would have an impact when operating on a larger scale. As Figure~\ref{fig:sl} shows, the performance of our method improves as the model size increases.

\noindent\textbf{Performance on different data size.} We conducted experiments on different ratios of training data to verify the robustness of LACT.
\begin{figure}[t]
    \centering
    \includegraphics[width=1\columnwidth]{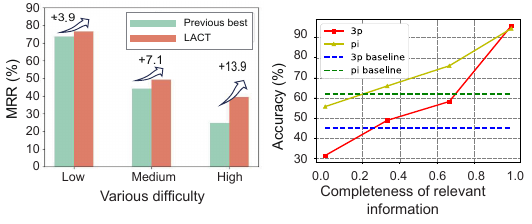}
    \caption{(left) \textbf{MRR} performance of LACT and previous SOTA methods at \textbf{different difficulties} based on FB15K. (right) The \textbf{correlation} between \textbf{related information completeness and accuracy} evaluated on FB15K-237, we selected a 3p query and pi query with the same inference path length as the task types. We assume that the completeness of all simple queries is 1.}
    \label{fig:2}
\end{figure}
\section{Discussion}

\subsection{When and Where Does LACT Work?}
The performance of LACT would be related to the following two aspects: $\textit{I}$. \textit{The completeness of relevant information extracted from KG. } $\textit{II}$. \textit{Sophistication of complex reasoning. }

\noindent\textbf{LACT performs consistently better with more complete information.} We take the form of a posteriori that set the completeness of relevant triplets to the proportion of triplets in the inference path of complex reasoning in the provided context, (For example, if the inference path of a 2i query is (Turing Award, winner, Yoshua Bengio), (Canada, citizen, Yoshua Bengio), and we can only retrieve (Turing Award, winner, Yoshua Bengio) in incomplete KG, then our completeness of relevant triplets to 1/2.) and set the completeness of simple queries that can be directly inferred to 1, to obtain the relation between Accuracy and correlation information completeness. As seen in Figure ~\ref{fig:2}, LACT obtains a significant gain when the completeness of relevant information increases, though, with zero relevant information, it remains a certain amount of complex reasoning ability.

\noindent\textbf{LACT performs consistently better on higher difficulties.} As mentioned before, we simply divide the difficulty of the query by the number of hops in the query. The results in Figure ~\ref{fig:2} show that our model yields more gain in tasks of higher-level difficulty and complexity, which benefits from our unique and sophisticated fine-tuning framework.

\section{Conclusion}

In this paper, we present a simple and effective fine-tuning framework LACT to boost the complex logical reasoning ability over KGs. LACT is a two-part method that consists of both Binary Tree Decomposition and Curriculum Learning and can be applied to various size LLMs with different data sizes. We empirically demonstrate the effectiveness and universality of the LACT on a series of widely used knowledge graph datasets. Further analyses delve into the underlying mechanism of our LACT, and investigate When and Why Does LACT Work. We hope that our work can inspire more research on combining knowledge graphs and LLMs.

\clearpage

\bibliography{aaai25}
\section{Acknowledgments}
This research is supported by NSFC (62225113, 62206202), National Key RD Program of China (2023YFC2705700), The Innovative Research Group Project of Hubei Province under Grants (2024AFA017), the National Research Foundation, Singapore, and the CyberSG R\&D Programme Office (“CRPO”), under the National Cybersecurity R\&D Programme (“NCRP”), RIE2025 NCRP Funding Initiative (Award CRPO-GC1-NTU-002).
\clearpage
\appendix
\begin{center}
\large
\textbf{Appendix}
\end{center}
\section{Neighborhood Retrieval Algorithm}\label{sec:neighborhood}
\label{NRA}
\subsection{Retrieval Algorithm}
To strike a balance between the completeness of relevant information and the token number limit of LLMs, we search for as many relevant triplets as possible along the possible paths.

Particularly, for the 1p query, we simply find all the triplets containing the entity or the relation.

\begin{figure}[ht]
    \centering
    \includegraphics[width=0.8\linewidth]{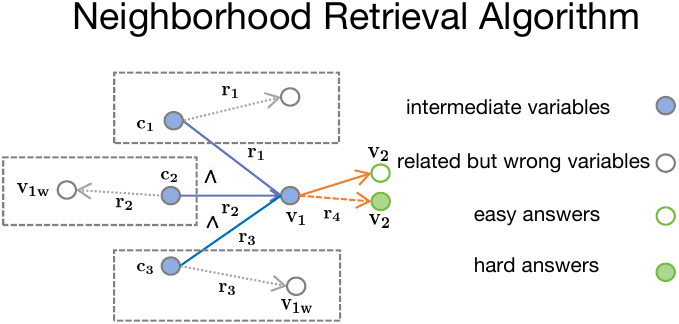}
    \caption{A case on Neighborhood Retrieval Algorithm}
    \label{fig:ir}
\end{figure}

For another query, as shown in Figure \ref{fig:ir} for each leaf node in DAG, we do depth traversal on the graph. For each step in the traversal process, if this step is a projection, we search for all the possible triplets. Otherwise, we perform corresponding operations on intersection and union respectively to filter out the corresponding entities.

We continue this traversal until the obtained entity is empty or reaches the root node. All triplets during the traversal are related to triplets. 
\subsection{Over-limit Solutions and Token distribution }
\begin{figure}[ht]
    \centering
    \includegraphics[width=0.8\linewidth]{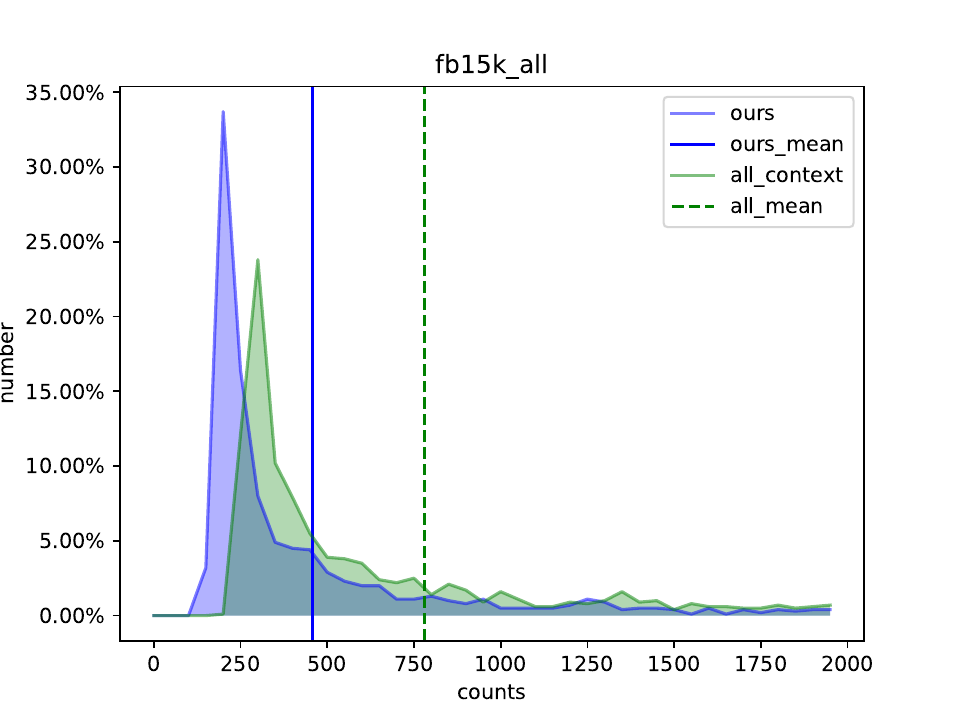}
    \caption{Token numbers' probability distribution in the dataset FB15K-237 and all-context Retrieval Algorithm.}
    \label{fig:token}
\end{figure}

In all experimental setups, we used the Max Seq Length of 4096, and we simply discarded all out-of-bounds samples and recorded them as 0 at the time of evaluation. In fact, after using our information retrieval algorithm, most of our samples were controlled below 4096. As shown in Figure \ref{fig:token}, compared with the algorithm of searching all possible related triples, our retrieval algorithm greatly reduces the consumption of tokens.
\begin{figure*}
    \centering
    \includegraphics[width=0.7\textwidth]{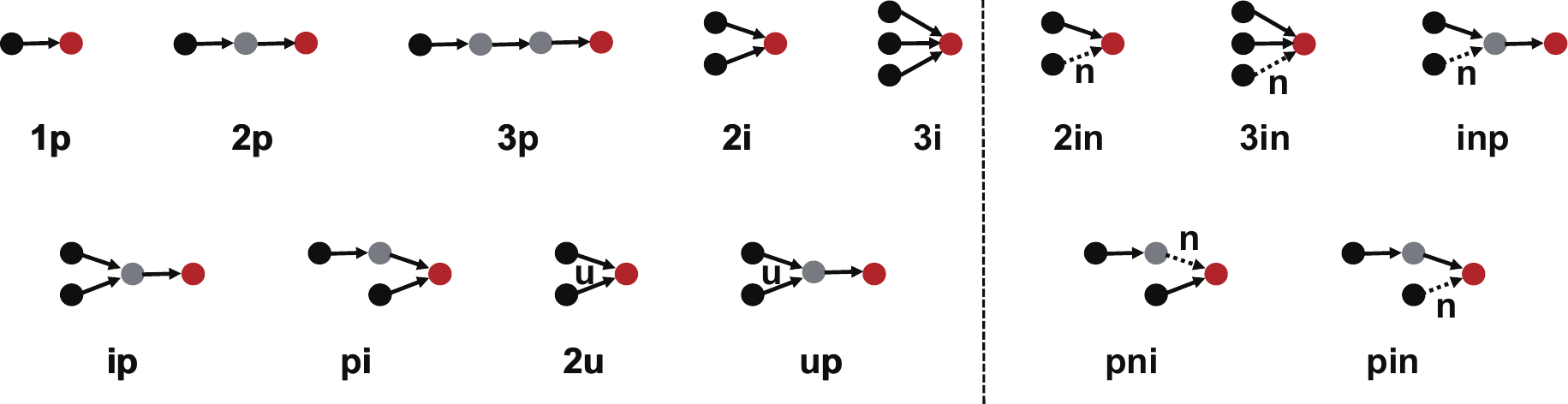}
    \caption{Query structure diagram of 14 types of queries}
    \label{fig:querystructure}
\end{figure*}

\begin{table*}
    \centering
    \begin{adjustbox}{width=0.95\textwidth}
    \begin{tabular}{lcccccccccccccc}
    \toprule
    & \bf{1p} & \bf{2p} & \bf{3p} & \bf{2i} & \bf{3i} & \bf{pi} & \bf{ip} & \bf{2u} & \bf{up} & \bf{2in} & \bf{3in} & \bf{inp} & \bf{pin} & \bf{pni} \\
    \midrule
    Number of subqueries &  1& 2 & 3&3 & 5 & 4 & 4  & 3 & 4 & 3 & 5 & 4 & 4& 4\\
    Difficulty&  1 &1&2&2 & 3& 3 & 3 & 2& 3 & 2 & 3 & 3 & 3 &3 \\
    \toprule
    \end{tabular}
    \end{adjustbox}
    \caption{Difficulty of different query types where 1 means easy, 2 means medium, 3 means hard.}
    \label{tb:Difficulty}
\end{table*}

\section{Universal Procedure for Converting from $EFO_{1}$ to Computational Trees}\label{sec:conversion}
To transform a $EFO_{1}$ formula into its corresponding computation tree, a two-phase process is employed: dependency graph generation, and union branches duplication.

\noindent\textbf{Dependency Graph Generation.} Upon encountering a $EFO_{1}$ expression, our primary procedure entails the allocation of distinct nodes to individual variables, while a separate node is assigned to each specific entity for each single-hop query. It is important to acknowledge that multiple nodes may represent one same intermediate entity, given its occurrence among various single-hop atom. Subsequently, undirected edges are employed to establish connections between nodes in accordance with the defined one-hop atoms. In the case where $e^i_j=r(h, t)$, we link $h$  and $t$ by establishing the edge recorded as $h_k$.
Similarly, if $e^i_j=\lnot r(h, t)$, we link $h$ and $t$ by the edge recorded as $\lnot r_k$.
The variable $k$ serves as a distinguishing label for edges emanating from distinct atom struture. The formulated  multi-graph must conform be in a tree-shape, signifying a DAG. Choosing tail node $v_?$ to be the root, We define the orientation of edges to consistently direct from child nodes towards their parent nodes, while carefully accommodating inverse relationships. It is important to note that constant entities nherently serve as terminal nodes in the computation tree, as they are each uniquely associated with a solitary variable node.

\noindent\textbf{Union Branches Duplication}
Subsequently we address the duplication branches within the computation tree.
For each path $\tau$ from every leaf node to root, we search for the first node $v_i$ which contains the same relations between $v_i$ and its possible child nodes $v_i^{'}$, but in different link structure: $r_{i_1}, r_{i_2}, \dots, r_{i_n}$.
These edges were amalgamate into a converged edge $r_{i_1, i_2, \dots i_n}$ to dissolve the complex multiple edges structure and then they can be amalgamate abide by the following distributive law:
\begin{equation}
    (A\land B)\lor(A\land C) \Leftrightarrow A\land(B\lor C)
\end{equation}
\section{Prompt Template}\label{sec:prompts}
The query and output prompt templates can be found in \ref{fig:query} and \ref{fig:answer}.
\begin{prompt}
    {\small 
    Prompt 1: Query Prompt Template of LACT.}
    \centering
    \includegraphics[width=1\linewidth]{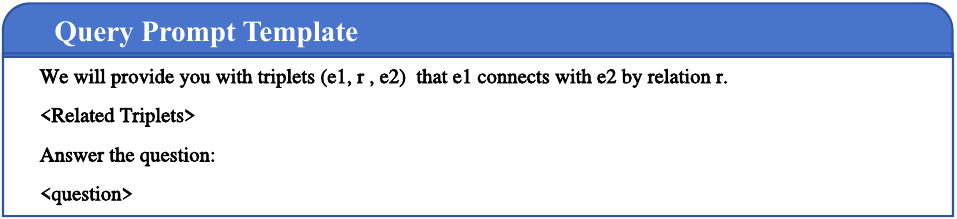}
    \label{fig:query}
\end{prompt}
\begin{prompt}
    \centering
    {
    \small 
    Prompt 2: Answer Template of LACT.
    }
    \includegraphics[width=1\linewidth]{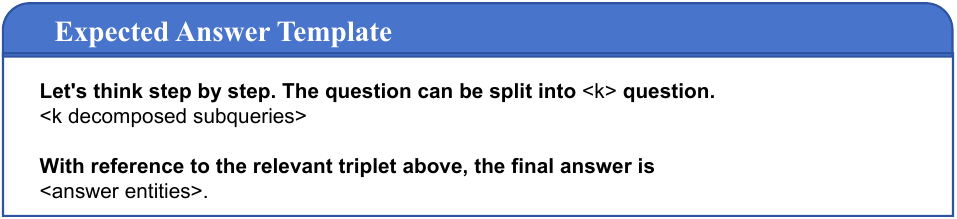}
    \label{fig:answer}
\end{prompt}

\section{Difficulty}

We divide the difficulty by the number of decomposed subqueries. Query types and their corresponding difficulties are shown in Table~\ref{tb:Difficulty}.

\section{Experiment Details}
\subsection{Dataset Details}\label{sec:dataset}
$\bullet$ \textbf{FB15K} comes from a subset of Freebase, a large knowledge database initiated by Google. FB15k has a total of 592,213 triplets consisting of 15,000 entities and 1,345 relationships.

\noindent$\bullet$ \textbf{FB15K-237} is the subset of FB15k which contains 310,116 triplets consisting of 14,541 entities and 237 relations. FB15K-237 corrects the inverse relationship leakage problem in the test set to a certain extent by selecting a specific relationship subset in FB15K, thereby more realistically reflecting the effect of the model on complex reasoning on the knowledge graph.

\noindent$\bullet$ \textbf{NELL995} is automatically extracted and generated by Carnegie Mellon University's Never-Ending Language Learning (NELL) project. It was considered as a comprehensive knowledge graph data set that includes 995 relationships in multiple fields.

\subsection{Training Details}  We use open-source model LLaMA-2-base, including two different parameter sizes: 7B and 13B, as the base model for fine-tuning. All LLaMA-2-7B and LLaMA-2-13B models are trained by fully fine-tuning. 

\noindent For the fully fine-tuning setting, we use the AdamW optimizer to train the model with 1 epoch and the batch size is 128. We use 8 NVIDIA A100 GPUS for training with the learning rate of 3e-6. 

\subsection{Training Cost}  We trained for a total of approximately 10 hours using 8*A100 during three training stages. 
\subsection{Query Structure}

Taking into account the fairness of the evaluation, we use the same 14 types of queries as in the previous work~\cite{bai2023answering}. The query structure of each type is shown in Figure~\ref{fig:querystructure}.

\begin{figure}[ht]
    \centering
    \includegraphics[width=0.7\columnwidth]{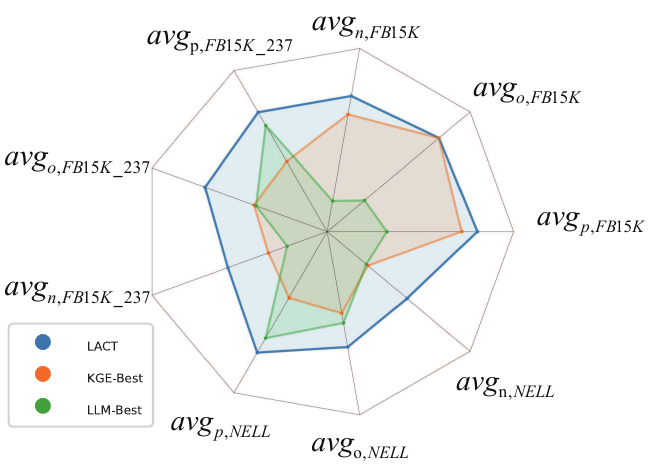}
    \caption{The results of the \textbf{main experiment}. We evaluate the performance of \textbf{three current state-of-the-art methods} on three datasets.}
    \label{fig:1}
\end{figure}
\begin{table*}[htbp]
    \centering
    \begin{adjustbox}{width=0.95\textwidth}
    \begin{tabular}{lcccccccccccccc}
    \toprule
    \bf{Method}  & \bf{1p} & \bf{2p} & \bf{3p} & \bf{2i} & \bf{3i} & \bf{pi} & \bf{ip} & \bf{2u} & \bf{up} & \bf{2in} & \bf{3in} & \bf{inp} & \bf{pin} & \bf{pni} \\
    \midrule
    Llama2-7B &  76.5 & 54.3 & 30.3 & 56.0 & 54.5 & 54.6&36.9 &56.5 &29.7&17.6&33.1&27.1&19.8&11.2 \\
    Mistral-7B&   \textbf{82.6}& \textbf{59.5} & 34.2& 59.6& \textbf{58.5} & 59.1&\textbf{39.1} &59.2 &\textbf{33.7}&20.4&39.2&\textbf{31.4}&\textbf{22.4}&\textbf{13.5}\\
    Qwen1.5-7B & 81.3& 57.2 & \textbf{36.8} & \textbf{62.9} & 58.1 & \textbf{63.9} & 38.6 & \textbf{60.4} & 31.8 & \textbf{20.7} &\textbf{ 41.9 }&28.7 & 20.8 & 11.6 \\
    \toprule
    \end{tabular}
    \end{adjustbox}
    \caption{\textbf{MRR} results (\%) based on
    differnet base models evaluated on FB15k-237.}
    \label{tb:ab3}
\end{table*}
\subsection{Evaluation Protocol Detail}\label{sec:eval protocal}

The answers to queries in complex logical reasoning can be divided into simple and hard types, which are distinguished by whether answers of the query can be easily retrieved directly in the knowledge graph. Specifically, in the valid set, we consider answers that can be directly retrieved from the graph in the training set as simple answers, and those that cannot be directly retrieved as hard answers. And in the test set, we consider answers that can be directly retrieved from the valid graph as simple answers, and those that cannot be directly retrieved as hard answers. Referring to previous work~\cite{Ren_Leskovec_2020}, we choose the Mean Reciprocal Rank (MRR) as the evaluation index on the filtered setting all answers (including easy and hard) would be filtered out when ranking.
\begin{figure}[htbp]
    \centering
    \includegraphics[width=0.8\columnwidth]{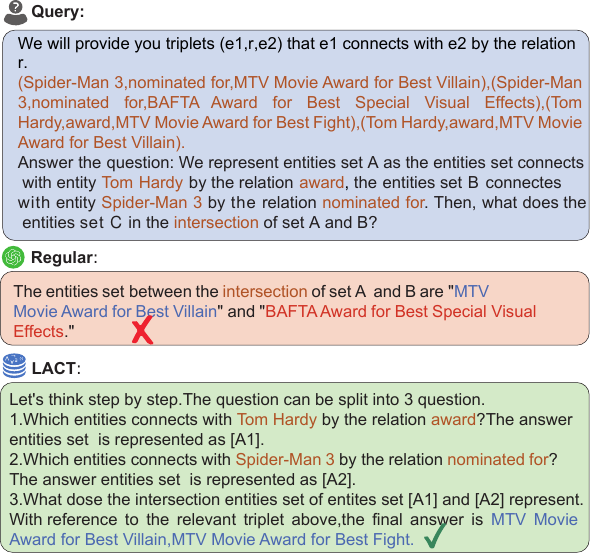}
    \caption{\textbf{Inference results} of ChatGPT and LACT on 2i query case, respectively.}
    \label{fig:case study}
\end{figure}
\begin{figure}[htbp]
    \centering
    \includegraphics[width=0.65\columnwidth]{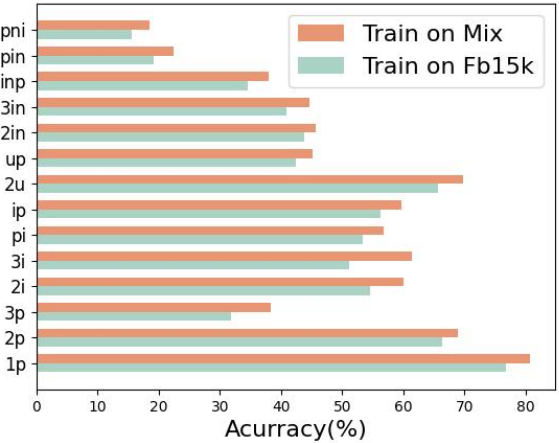}
    \caption{Ablation experimental results of Accuracy (\%) \textbf{trained on FB15k} and \textbf{tested on FB15K-237}, compared to models \textbf{trained on all mixed training data}.}
    \label{fig:ablation}
\end{figure}
\section{More Results}
\subsection{Overall comparison}
We provide a multi-dimensional comparison with previous state-of-the-art methods in Figure ~\ref{fig:1}
\subsection{OOD results}
\begin{table*}[htbp]
    \centering
    \begin{adjustbox}{width=0.95\textwidth}
    \begin{tabular}{lcccccccccccccc}
    \toprule
    & \bf{1p} & \bf{2p} & \bf{3p} & \bf{2i} & \bf{3i} & \bf{pi} & \bf{ip} & \bf{2u} & \bf{up} & \bf{2in} & \bf{3in} & \bf{inp} & \bf{pin} & \bf{pni} \\
    \midrule
    \textbf{LARK}  & 72.8 & 50.7 & 36.2 & 66.9 & 60.4 & 56.1 & 23.5 & 52.4 & 40.6& 16.2 & 5.7 &33.7 & 26.1& 10.0\\
   \textbf{LACT}& \textbf{97.0}&\textbf{87.2}&\textbf{80.8}&\textbf{95.8}& \textbf{85.3}&\textbf{ 71.7} & \textbf{77.4}& \textbf{85.7}& \textbf{62.6} & \textbf{88.5} & \textbf{81.4} & \textbf{77.8} & \textbf{68.9} &\textbf{65.7} \\
    \toprule
    \end{tabular}
    \end{adjustbox}
    \caption{Accuracy results (\%) evaluated on \textbf{FB15k} on \textbf{complete KG} for complex query on all query types.}
    \label{tb:allcontext}
\end{table*}
\begin{table}[ht]
    \centering
    \begin{adjustbox}{width=0.45\textwidth}
    \begin{tabular}{lcccccc}
    \toprule
    \bf{Metric} &  \bf{pi} & \bf{ip} & \bf{2u} & \bf{up} &\bf{nin} &\bf{nipn} \\
    \midrule
    Pro$_{decomposed}$ & 98.7 & 100.0 & 97.8 & 100.0 & 98.4 &97.2\\
    Pro$_{true,decomposed}$&  98.6 & 99.9 &97.8&99.6 &96.1 &96.9\\
    \toprule
    \end{tabular}
    \end{adjustbox}
    \caption{In \textbf{OOD} queries, the proportion of queries that can be \textbf{decomposed} and the proportion of queries that can be \textbf{decomposed correctly} on FB15k.}
    \label{tb:ab4}
\end{table}

As shown in Table~\ref{tb:ab4}including pi/ip/2u/up more than 95\% of OOD test samples can still follow logical chain reasoning. These phenomena indicate strong generalization ability of LACT. In fact, many query types including nin/nipn that have not been tested before can also be correctly decomposed and queried by LACT.
\subsection{Dataset Ablation Results}

The results in Figure~\ref{fig:ablation} show that the reasoning ability stimulated by one dataset can still be demonstrated in another dataset, which reflects well in the query performance which only dropped less than 5\%.

\subsection{Model-level transferability}
The results in Table ~\ref{tb:ab3} show that LACT is effective on different base models and benefits from the improvement of the base model.
\subsection{Case Study}
To have a close and interen look, we perform the case studies by analyzing the results of LACT and ChatGPT~(GPT-3.5-turbo-0613). As shown in Figure~\ref{fig:case study}, ChatGPT cannot make good use of incomplete knowledge graphs for reasoning in some cases. Conversely, LACT performs reasoning through a complete logical chain, making maximum use of the relevant information provided and deducing the correct answer, which greatly improves the reasoning ability.

\end{document}